\documentclass{article}
\usepackage{times}
\usepackage{graphicx} 
\usepackage{subfigure} 
\usepackage{natbib}
\usepackage{algorithm}
\usepackage{algorithmic}

\usepackage{helvet}
\usepackage{courier}
\usepackage{tikz}
\usepackage{amsmath}
\usepackage{amsthm}
\usepackage{amssymb}
\usepackage{graphicx} 
\usepackage{subfigure}
\usepackage{enumerate}
\usepackage{bm}
\usepackage{multirow}
\usepackage{wrapfig}

\usepackage{caption}

\newcommand{\RN}[1]{%
	\textup{\lowercase\expandafter{\it \romannumeral#1}}%
}

\newcommand{\ie}{\textit{i.e.}}
\newcommand{\eg}{\textit{e.g.}}

\usepackage[accepted]{icml2017}

\usepackage[utf8]{inputenc} 
\usepackage[T1]{fontenc}    
\usepackage{hyperref}       
\usepackage{url}            
\usepackage{booktabs}       
\usepackage{amsfonts}       
\usepackage{nicefrac}       
\usepackage{microtype}      

\usepackage{amsmath}
\usepackage{amsfonts}
\usepackage{amssymb}
\usepackage{wrapfig}
\usepackage{bm}
\usepackage{psfrag}
\usepackage{tikz}
\usepackage{epstopdf}

\usepackage{color}
\usepackage{tikz}
\usetikzlibrary{arrows,shapes,snakes,automata,backgrounds,fit,petri}
\usepackage{adjustbox}

\usepackage{algorithm}
\usepackage{algorithmic}

\newcommand{\beq}{\vspace{0mm}\begin{equation}}
\newcommand{\eeq}{\vspace{0mm}\end{equation}}
\newcommand{\beqs}{\vspace{0mm}\begin{eqnarray}}
\newcommand{\eeqs}{\vspace{0mm}\end{eqnarray}}
\newcommand{\barr}{\begin{array}}
\newcommand{\earr}{\end{array}}

\newcommand{\Cmat}{{\bf C}}

\newcommand{\Imat}{{\bf I}}

\newcommand{\Vmat}[0]{{{\bf V}}}
\newcommand{\Wmat}[0]{{{\bf W}}}
\newcommand{\Xmat}[0]{{{\bf X}}}

\newcommand{\bv}[0]{{\boldsymbol{b}}}
\newcommand{\cv}[0]{{\boldsymbol{c}}}

\newcommand{\fv}[0]{{\boldsymbol{f}}}

\newcommand{\hv}[0]{{\boldsymbol{h}}}

\newcommand{\xv}{\boldsymbol{x}}
\newcommand{\yv}{\boldsymbol{y}}
\newcommand{\zv}{\boldsymbol{z}}

\newcommand{\Sigmamat}[0]{{\boldsymbol{\Sigma}}}

\newcommand{\muv}[0]{{\boldsymbol{\mu}}}

\newcommand{\phiv}{\boldsymbol{\phi}}

\newcommand{\R}{\mathbb{R}}

\newcommand{\Xcal}{\mathcal{X}}
\newcommand{\Ycal}{\mathcal{Y}}

\newcommand{\Lcal}{\mathcal{L}}

\newcommand{\Scal}{\mathcal{S}}

\newcommand{\Ucal}{\mathcal{U}}
\newcommand{\Ocal}{\mathcal{O}}

\icmltitlerunning{Adversarial Feature Matching for Text Generation}

\begin{document} 

\twocolumn[
\icmltitle{Adversarial Feature Matching for Text Generation}




\begin{icmlauthorlist}
\icmlauthor{Yizhe Zhang}{duke}
\icmlauthor{Zhe Gan}{duke}
\icmlauthor{Kai Fan}{duke}
\icmlauthor{Zhi Chen}{duke}
\icmlauthor{Ricardo Henao}{duke}
\icmlauthor{Dinghan Shen}{duke}
\icmlauthor{Lawrence Carin}{duke}
\end{icmlauthorlist}

\icmlaffiliation{duke}{Duke University, Durham, NC, 27708}
\icmlcorrespondingauthor{Yizhe Zhang}{yizhe.zhang@duke.edu}
\icmlkeywords{GAN, generative adversarial network, deep learning}

\vskip 0.3in
]



\printAffiliationsAndNotice{}  


\begin{abstract}
	The Generative Adversarial Network (GAN) has achieved great success in generating realistic (real-valued) synthetic data. However, convergence issues and difficulties dealing with discrete data hinder the applicability of GAN to text. We propose a framework for generating realistic text via adversarial training. We employ a long short-term memory network as generator, and a convolutional network as discriminator.
	Instead of using the standard objective of GAN, we propose matching the high-dimensional latent feature distributions of real and synthetic sentences, via a kernelized discrepancy metric.
	This eases adversarial training by alleviating the mode-collapsing problem.
	Our experiments show superior performance in quantitative evaluation, and demonstrate that our model can generate realistic-looking sentences.
\end{abstract}

\section{Introduction}
Generating meaningful and coherent sentences is central to many natural language processing applications.
The general idea is to estimate a distribution over sentences from a corpus, then use it to sample realistic-looking sentences.
This task is important because it enables generation of \emph{novel} sentences that preserve the semantic and syntactic properties of real-world sentences, while being potentially different from any of the examples used to estimate the model.
For instance, in the context of dialog generation, it is desirable to generate answers that are more diverse and less generic \cite{li2016deep}.

One simple approach consists of first learning a latent space to represent (fixed-length) sentences using an encoder-decoder (autoencoder) framework based on Recurrent Neural Networks (RNNs) \cite{cho2014learning,sutskever2014sequence}, then generate synthetic sentences by decoding random samples from this latent space.
However, this approach often fails to generate realistic sentences from arbitrary latent representations.
The reason for this is that, when mapping sentences to their latent representations using an autoencoder, the mappings usually cover a small but structured region of the latent space, which corresponds to a manifold embedding \cite{bowman2015generating}.
In practice, most regions of the latent space do not necessarily map (decode) to realistic sentences.
Consequently, randomly sampling latent representations often yields nonsensical sentences.
Recent work by \citet{bowman2015generating} has attempted to generate more diverse sentences via RNN-based variational autoencoders.
However, they did not address the fundamental problem that the posterior distribution over latent variables does not appropriately cover the latent space.

Another underlying challenge of generating realistic text relates to the nature of the RNN.
During inference, the RNN generates words in sequence from previously {\em generated} words, contrary to learning, where ground-truth words are used every time.
As a result, error accumulates proportional to the length of the sequence, \ie, the first few words look reasonable, however, quality deteriorates quickly as the sentence progresses.
\citet{bengio2015scheduled} coined this phenomenon \emph{exposure bias}.
Toward addressing this problem, \citet{bengio2015scheduled} proposed the scheduled sampling approach.
However, \citet{huszar2015not} showed that scheduled sampling is a fundamentally inconsistent training strategy, in that it produces largely unstable results in practice.

The Generative Adversarial Network (GAN) \cite{goodfellow2014generative} is an appealing and natural answer to the above issues.
GAN matches the distributions of synthetic and real data by introducing an adversarial game between a \emph{generator} and a \emph{discriminator}.
The GAN objective seeks to constitute a generator, that functionally maps samples from a given (simple) prior distribution, to synthetic data that appear to be \emph{realistic}.
The GAN setup explicitly seeks that the latent representations from real data (via encoding) be distributed in a manner consistent with the specified prior (\eg, Gaussian or uniform).
Due to the nature of adversarial training, the discriminator compares real and synthetic sentences, rather than their individual words, which in principle should alleviate the exposure-bias issue. Recent work \cite{lamb2016professor} has incorporated an additional discriminator to train a sequence-to-sequence language model that better preserves long-term dependencies. 

Effort has also been made to generate realistic-looking sentences via adversarial training.
For instance, by borrowing ideas from reinforcement learning, \citet{yu2016sequence,li2017adversarial} treat the sentence generation as a sequential decision making process.
Despite the success of these methods, two fundamental problems of the GAN framework limit their use in practice:
(\emph{i}) the generator tends to produce a single observation for multiple latent representations, \ie, mode collapsing \cite{metz2016unrolled}, and
(\emph{ii}) the generator's contribution to the \emph{learning signal} is insubstantial when the discriminator is close to its local optimum, \ie, vanishing gradient behavior \cite{arjovsky2017towards}.

In this paper we propose a new framework, \emph{TextGAN}, to alleviate the problems associated with generating realistic-looking sentences via GAN.
Specifically, the Long Short-Term Memory (LSTM) \cite{hochreiter1997long} RNN is used as generator, and the Convolutional Neural Network (CNN)~\cite{kim2014convolutional} is used as discriminator.
We consider a kernel-based moment-matching scheme over a Reproducing Kernel Hilbert Space (RKHS), to force the empirical distributions of real and synthetic sentences to have matched moments in latent-feature space.
As a consequence, our approach ameliorates the mode-collapsing issue associated with standard GAN training.
This strategy encourages the model to learn representations that are both informative of the original sentences (via the autoencoder) and discriminative w.r.t. synthetic sentences (via the discriminator).
We also propose several complementary techniques, including initialization strategies and discretization approximations to ease GAN training, and to achieve superior performance compared to related approaches.

\section{Model}\label{sec:model}
\subsection{Generative Adversarial Networks}
GAN \cite{goodfellow2014generative} aims to obtain the equilibrium of the following optimization objective
\begin{align}\label{eq:gan}
\hspace{-2.5mm}\Lcal_{GAN} = \mathbb{E}_{x\sim p_x} \log D(x) + \mathbb{E}_{z\sim p_z} \log [1-D(G(z))] ,
\end{align}
where $\Lcal_{GAN}$ is maximized w.r.t. $D(\cdot)$ and minimized w.r.t. $G(\cdot)$.
Note that the first term of \eqref{eq:gan} does not depend on $G(\cdot)$.
Observed (real) data, $x$, are sampled from empirical distribution $p_x(\cdot)$.
The latent code, $z$, that feeds into the generator, $G(z)$, is drawn from a simple prior distribution $p_z(\cdot)$.
When the discriminator is \emph{optimal}, solving this adversarial game is equivalent to minimizing the Jenson-Shannon Divergence (JSD) \cite{arjovsky2017towards} between the real data distribution $p_x(\cdot)$ and the synthetic data distribution $p_{\tilde{x}}(\cdot) \triangleq p(G(z))$ , where $z\sim p_z(\cdot)$ \cite{goodfellow2014generative}.
However, in most cases, the saddle-point solution of the objective in \eqref{eq:gan} is intractable.
Therefore, a procedure to iteratively update $D(\cdot)$ and $G(\cdot)$ is often applied.

\citet{arjovsky2017towards} pointed out that the standard GAN objective in \eqref{eq:gan} suffers from an unstably weak learning signal when the discriminator gets close to local optimal, due to the gradient-vanishing effect.
This is because the JSD implied by the original GAN loss becomes a constant if $p_x(\cdot)$ and $p_{\tilde{x}}(\cdot)$ share no support, thus minimizing the JSD yields no learning signal.
This problem also exists in the recently proposed energy-based GAN (EBGAN) \cite{zhao:2016wu}, as the distance metric implied by EBGAN is the Total Variance Distance (TVD), which has the same issue w.r.t. JSD, as shown by \citet{arjovsky2017wasserstein}.

\subsection{TextGAN}
\begin{figure}[t!]
	\centering
	\includegraphics[width=.35\textwidth]{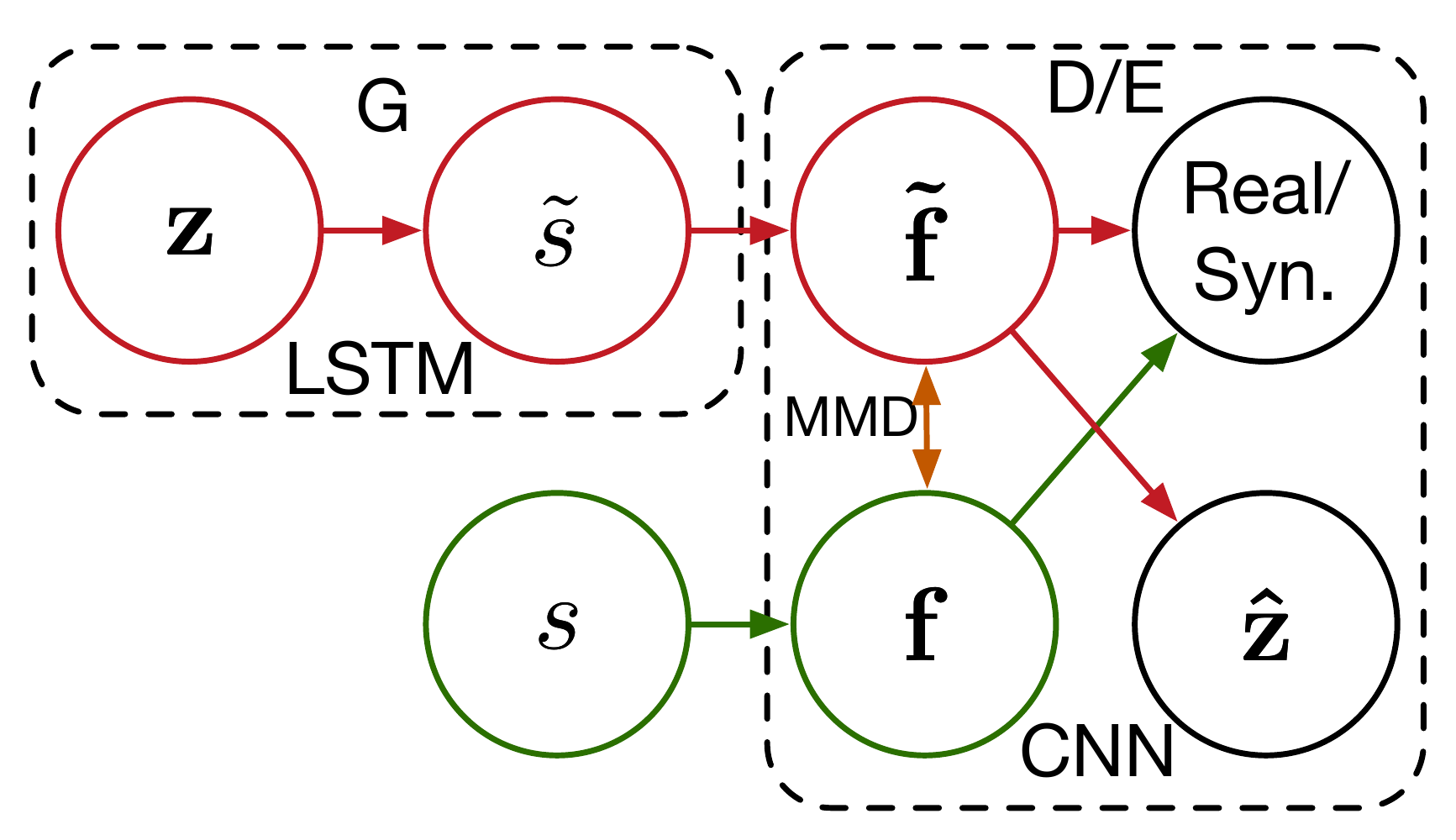}
	\caption{Model scheme of TextGAN. Latent codes $z$ are fed through a generator $G(\cdot)$, to produce synthetic sentence $\tilde{s}$. Synthetic and real sentences ($\tilde{s}$ and $s$) are fed into a binary discriminator $D(\cdot)$, for real \emph{vs.} fake (synthetic) prediction, and also for latent code reconstruction $\hat{z}$. $\tilde{\fv}$ and $\fv$ represent features of $\tilde{s}$ and $s$, respectively.}
	\label{fig:textGAN}
	\vspace{-3mm}
\end{figure}

Given a sentence corpus $\Scal$, instead of directly optimizing the objective from standard GAN in \eqref{eq:gan}, we adopt an approach that is similar to the feature matching scheme of \citet{salimans2016improved}.
Specifically, we consider the objective
\begin{align}
\Lcal_D = & \ \Lcal_{GAN} - \lambda_{r} \Lcal_{recon} + \lambda_{m} \Lcal_{MMD^2}   \label{eq:d_step_mmd} \\
\Lcal_G = & \ \Lcal_{MMD^2} \label{eq:g_step_mmd} \\
\Lcal_{GAN} = & \ \mathbb{E}_{s\sim \Scal} \log D(s) + \mathbb{E}_{z\sim p_z} \log [1-D(G(z))] \nonumber \\
\Lcal_{recon} = & \ ||\hat{z} - z||^2 \,, \nonumber
\end{align}
where $\Lcal_D$ and $\Lcal_G$ are iteratively maximized w.r.t $D(\cdot)$ and minimized w.r.t. $G(\cdot)$, respectively.
$\Lcal_{GAN}$ is the standard objective of GAN in \eqref{eq:gan}.
$\Lcal_{recon}$ is the Euclidean distance between the reconstructed latent code, $\hat{z}$, and the original code, $z$, drawn from prior distribution $p_z(\cdot)$.
We denote the synthetic sentences as $\tilde{s} \triangleq G(z)$, where $z \sim p_z(\cdot)$.
$\Lcal_{MMD^2}$ represents the Maximum Mean Discrepancy (MMD) \cite{gretton2012kernel} between the empirical distribution of sentence embeddings $\tilde{\fv}$ and $\fv$, for synthetic and real data, respectively.
The model framework is illustrated in Figure~\ref{fig:textGAN} and detailed below.

We first consider $\Lcal_G$ in \eqref{eq:g_step_mmd}.
The generator $G(\cdot)$ attempts to adjust itself to produce synthetic sentence $\tilde{s}$, with
features $\tilde{\fv}$, encoded by $D(\cdot)$, to mimic the real sentence features $\fv$ (also encoded by $D(\cdot)$).
This is achieved by matching the empirical distributions of $\tilde{\fv}$ and $\fv$ via the MMD objective. 

Concisely, MMD measures the mean squared difference between two sets of samples $\Xcal$ and $\Ycal$, where $\Xcal=\{x_i\}_{i=1:N_x}$, $x_i \in \mathbb{R}^d$, $\Ycal=\{y_i\}_{i=1:N_y}$, $y_i \in \mathbb{R}^d$, $d$ is the dimensionality of the samples, and $N_x$ and $N_y$ are sample sizes for $\Xcal$ and $\Ycal$, respectively.
The MMD metric characterizes the differences between $\Xcal$ and $\Ycal$ over a Reproducing Kernel Hilbert Space (RKHS), $\mathcal{H}$, associated with kernel function $k(\cdot): \mathbb{R}^d \times \mathbb{R}^d \mapsto \mathbb{R}$.
The kernel can be written as an inner product over $\mathcal{H}$: $k(x,x')= \langle k(x,\cdot),k(x',\cdot) \rangle_{\mathcal{H}}$, and $\phiv(x) \triangleq k(x,\cdot) \in \mathcal{H}$ is denoted as the \emph{feature mapping} \cite{gretton2012kernel}.
Formally, the MMD for two empirical distributions $\Xcal$ and $\Ycal$ is given by
\begin{align}
\Lcal_{MMD^2} = & \ || \mathbb{E}_{x\sim \Xcal} \phiv(x) - \mathbb{E}_{y \sim \Ycal} \phiv(y) ||_\mathcal{H}^2 \label{eq:mmd}\\
= & \ \mathbb{E}_{x\sim \Xcal} \mathbb{E}_{x'\sim \Xcal} [k(x,x')] \nonumber \\
+ & \ \mathbb{E}_{y\sim \Ycal} \mathbb{E}_{y'\sim \Ycal} [k(y,y')]
- 2 \mathbb{E}_{x\sim \Xcal} \mathbb{E}_{y\sim \Ycal} [k(x,y)] . \nonumber
\end{align}
Note that $\Lcal_{MMD^2}$ reaches its minimum when the two empirical distributions $\Xcal$ and $\Ycal$ (in general) match exactly.
For example, with a polynomial kernel, $k(x,y)=(x^T y + c)^L$, minimizing $\Lcal_{MMD^2}$ can be understood as matching moments of two empirical distributions up to order $L$.
With a universal kernel like the Gaussian kernel, $k(x,y) = \exp(- \frac {||x-y||^2} {2 \sigma} )$, with bandwidth $\sigma$, minimizing the MMD objective will match moments of all orders \cite{gretton2012kernel}.
Here, we use MMD to match the empirical distribution of $\tilde{\fv}$ and $\fv$ using a Gaussian kernel.

The adversarial discriminator $D(\cdot)$ associated with the loss in \eqref{eq:d_step_mmd} aims to produce sentence features that are most \emph{discriminative}, \emph{representative} and \emph{challenging}.
These aims are explicitly represented as the three components of \eqref{eq:d_step_mmd}, namely,
(\emph{i}) $\Lcal_{GAN}$ requires $\tilde{\fv}$ and $\fv$ to be discriminative of real and synthesized sentences;
(\emph{ii}) $\Lcal_{recon}$ requires $\tilde{\fv}$ and $\fv$ to preserve maximum reconstruction information for the latent code $z$ that generates synthetic sentences; and
(\emph{iii}) $\Lcal_{MMD^2}$ forces $D(\cdot)$ to select the most challenging features for the generator to match.

In the situation for which simple features are enough for the discrimination/reconstruction task, this additional loss seeks to estimate complex features that are difficult for the current generator, thus improving in terms of generation ability. In our experience, we find the reconstruction and MMD loss in D serve as regularizer to the binary classification loss, in that by adding these losses, discriminator features tend to be more spread-out in the feature space.

In summary, the adversarial game associated with \eqref{eq:d_step_mmd} and \eqref{eq:g_step_mmd} is the following: $D(\cdot)$ attempts to select informative sentence features, while $G(\cdot)$ aims to match these features.
Parameters $\lambda_r$ and $\lambda_m$ act as trade-off between discrimination ability, and reconstruction and moment matching precision, respectively.
We argue that this framework has several advantages over the standard GAN objective in \eqref{eq:gan}. 

The original GAN objective has been shown to be prone to mode collapsing, especially when the so-called $\log D$ alternative for the generator loss is used \cite{metz2016unrolled}, \ie, replacing the second term of \eqref{eq:gan} by $-\mathbb{E}_{z\sim p_z} \log [D(G(z))]$.
This is because when $\log D$ is used, fake-looking samples are penalized more severely than less diverse samples \cite{arjovsky2017towards}, thus grossly underestimating the variance of latent features.
The loss in \eqref{eq:g_step_mmd}, on the other hand, forces the generator to produce highly diverse sentences to match the variation of real sentences, by latent moment matching, thus alleviating the mode-collapsing problem. We believe that leveraging MMD is general enough to be useful as a framework in other data domains, \eg, images. Presumably, the discrete nature of text data makes standard GAN prone to mode-collapsing. This is manifested by close neighbors in latent code space producing the same text output. In our approach, MMD and feature matching are introduced to alleviate mode collapsing with text data as motivating domain.
However, whether such an objective is free from the convergence issues of the standard GAN, due to vanishing gradient from the generator, is known to be problem specific \cite{arjovsky2017towards}. 

\citet{arjovsky2017towards} demonstrated that JSD yields weak gradient signals when the real and synthetic data are far apart.
To deliver stable gradients, a smoother distance metric over the data domain is required.
In \eqref{eq:mmd}, we are essentially employing a Neural Network (NN) embedding via Gaussian kernel for matching $s$ and $\tilde{s}$, \ie, $k_s(s,s')=\phiv(g(s))^T \phiv(g(s'))$, where $g(\cdot)$ denotes the NN embedding that maps from the data to the feature domain.
Under the assumption that $g(\cdot)$ is a bijective mapping, \ie, distinct sentences have different embedded feature vectors, in the Supplementary Material we prove that if the original kernel function $k(x,y)= \phiv(x)^T \phiv(y)$ is universal, the composed kernel $k_s(s,s')$ is also universal.
As shown in \citet{gretton2012kernel}, the MMD is a proper metric when the kernel is universal.
In fact, if the kernel function is universal,
the MMD metric will be no worse than TVD in terms of vanishing gradients \cite{arjovsky2017wasserstein}.
However, if the bandwidth of the kernel is too small, much smaller than the average distance between data points, the vanishing gradient problem remains \cite{arjovsky2017wasserstein}.

Additionally, seeking to match the sentence features provides a more achievable and informative objective than directly trying to mislead the discriminator as in standard GAN.
Specifically, the loss in \eqref{eq:g_step_mmd} implies a clearer aim for the generator, as it requires matching the latent features (distribution-wise) as opposed to uniquely trying to fake a binary classifier. 

Note that if the latent features from real and synthetic data have similar distributions it is unlikely that the discriminator, that uses these features as inputs, will be able to tell them apart.
Implementation-wise, the updating signal from the generator does not need to propagate all the way back from the discriminator, but rather directly from the features layer, thus less prone to fading. We believe there may be other possible approaches for text generation using GAN, however, we hope to provide a first attempt toward overcoming some of the difficulties associated with it.

\subsection{Alternative (data efficient) objectives}\label{sec:alter}
One limitation of the proposed approach is that the dimensionality of features $\tilde{\fv}$ and $\fv$ could be much larger than the size of the subset of data (minibatch) used during learning, hence the empirical distribution may not be sufficiently representative.
In fact, a reliable Gaussian kernel MMD two-sample test generally requires the size of the minibatch to be proportional to the number of dimensions \cite{ramdas2014high}.
To alleviate this issue, we consider two strategies.

\paragraph{Compressing network} We map $\tilde{\fv}$ and $\fv$ into a lower-dimensional feature space using a \emph{compressing network} with fully connected layers, also learned by $D(\cdot)$.
This is sensible because the discriminator will still encourage the most challenging features to be abstracted (compressed) from the original features $\tilde{\fv}$ and $\fv$.
This approach provides significant computational savings, as computation of the MMD in \eqref{eq:mmd} scales with $\Ocal(d^2 d_f)$, where $d_f$ denotes the dimensionality of the feature vector.
However, a lower-dimensional mapping may miss valuable information.
Besides, finding the optimal mapping dimension may be difficult in practice.
There exists a tradeoff between fast estimation and a richer feature vector, by setting $d_f$ appropriately.

\paragraph{Gaussian covariance matching}
We could also avoid using the \emph{kernel trick}, as was used in \eqref{eq:mmd}. Instead, we can replace $\Lcal_{MMD^2}$ by $\Lcal_{G}^{(c)}$ (below), where we accumulate (Gaussian) sufficient statistics from multiple minibatches, thus alleviating the inadequate-minibatch-size issue. Specifically,
\begin{align}
\Lcal_G^{(c)} = & \ \text{tr}(\tilde{\Sigmamat}^{-1} \Sigmamat + \Sigmamat^{-1}\tilde{\Sigmamat}) \nonumber \\
+ & \ (\tilde{\muv} - \muv)^T (\tilde{\Sigmamat}^{-1} + \Sigmamat^{-1}) (\tilde{\muv} - \muv) \,, \label{eq:cov_match}
\end{align}
where $\tilde{\Sigmamat}$ and $\Sigmamat$ represent the covariance matrices of synthetic and real sentence feature vectors $\tilde{\fv}$ and $\fv$, respectively.
$\tilde{\muv}$ and $\muv$ denote the mean vectors of $\tilde{\fv}$ and $\fv$, respectively.
By setting $\tilde{\Sigmamat}=\Sigmamat=\Imat$, \eqref{eq:cov_match} reduces to the first-moment feature matching technique from \citet{salimans2016improved}.
Note that this loss $\Lcal_G^{(c)}$ is an upper bound of the JSD (omitting constant, proved in the Supplementary Material) between two multivariate Gaussian distribution $\mathcal{N}(\muv,\Sigmamat)$ and $\mathcal{N}(\tilde{\muv},\tilde{\Sigmamat})$, which is more tractable than directly minimizing JSD.
The feature vectors used in \eqref{eq:cov_match} are the neural net outputs before applying any non-linear activation function.
We note that the Gaussian assumption may still be strong in many cases.
In practice, we use a moving average of the most recent $m$ minibatches for estimating all sufficient statistics $\tilde{\Sigmamat}, \Sigmamat, \tilde{\muv}$ and $ \muv$.
Further, $\tilde{\Sigmamat}$ and $\Sigmamat$ are initialized to be $\Imat$ to prevent numerical problems.

\begin{figure}
	\centering
	\includegraphics[width=0.45\textwidth]{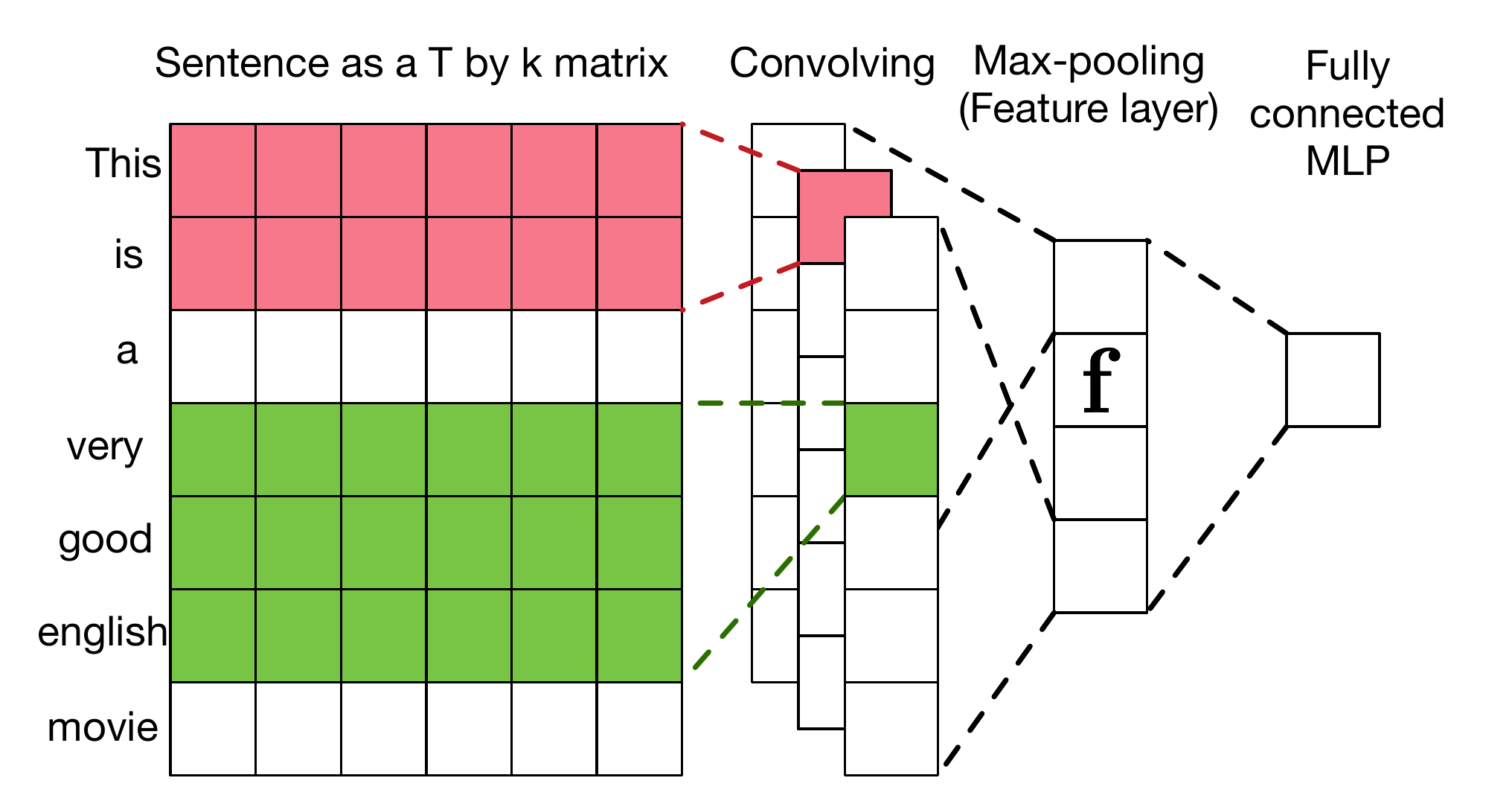}
	\includegraphics[width=0.45\textwidth]{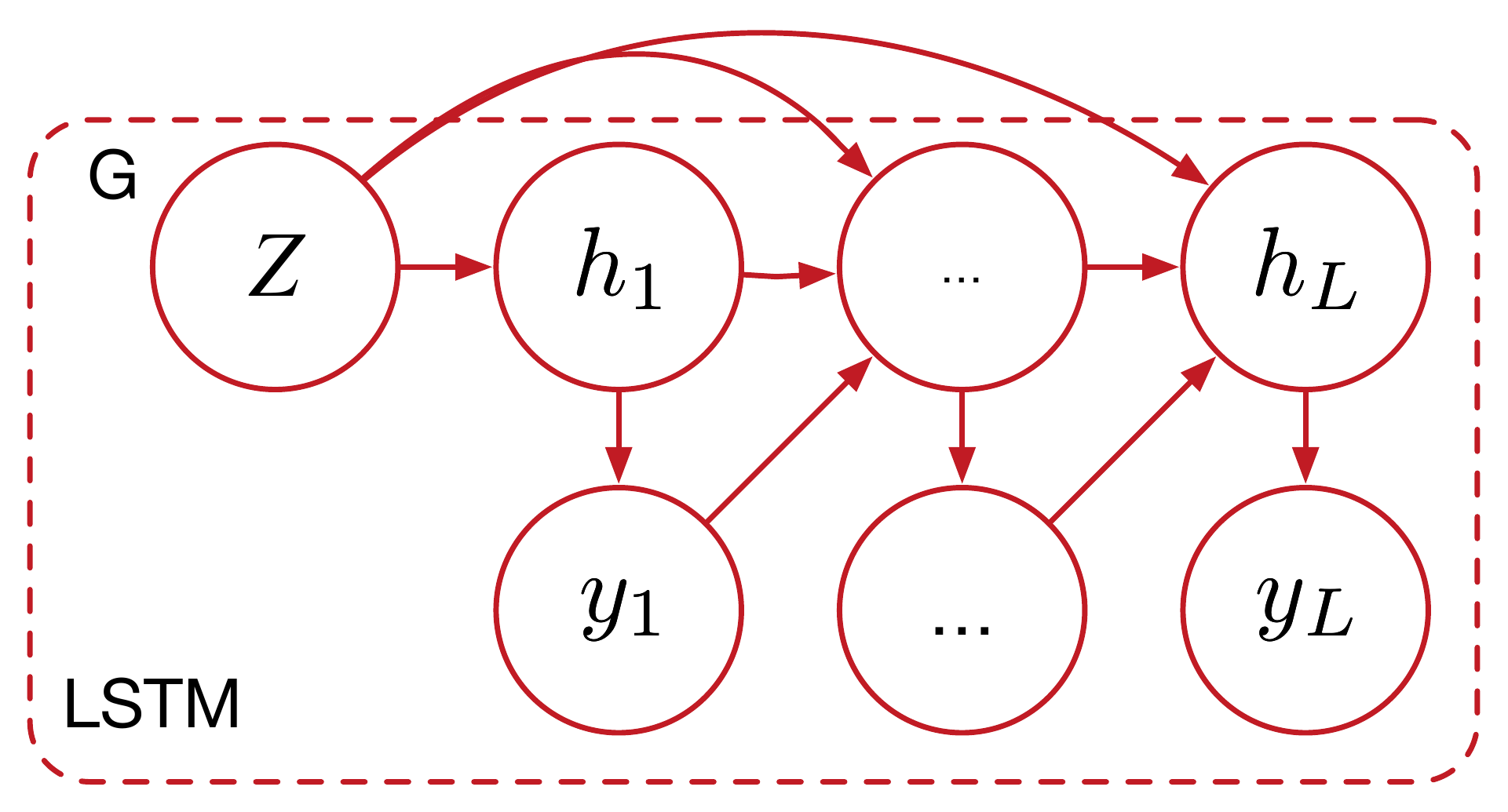}
	\caption{\small{Top: CNN-based sentence discriminator/encoder. Bottom: LSTM sentence generator.}}
	\label{fig:framework}
	\vspace{-3.0mm}
\end{figure}

\subsection{Model specification}
Let $w^t$ denote the $t$-th word in sentence $s$.
Each word $w^t$ is embedded into a $k$-dimensional word vector $\xv_t=\Wmat_e[w^t]$, where $\Wmat_e \in \R^{k\times V}$ is a (learned) word embedding matrix, $V$ is the vocabulary size, and notation $\Wmat_e[v]$ denotes the $v$-th column of matrix $\Wmat_e$.

\paragraph{CNN discriminator}
We use the CNN architecture in~\citet{kim2014convolutional,collobert2011natural} for sentence encoding.
It consists of a convolution layer and a max-pooling operation over the entire sentence for each feature map.
A sentence of length $T$ (padded where necessary) is represented as a matrix $\Xmat \in \R^{k\times T}$, by concatenating its word embeddings as columns, \emph{i.e.}, the $t$-th column of $\Xmat$ is $\xv_t$.

As shown in Figure~\ref{fig:framework}(top), a convolution operation involves a filter $\Wmat_c \in \R^{k\times h}$, applied to a window of $h$ words to produce a new feature.
Following \citet{collobert2011natural}, we induce a latent feature map $\cv=\gamma(\Xmat \ast \Wmat_c +\bv) \in \R^{T-h+1}$, where $\gamma(\cdot)$ is a nonlinear activation function (we use the hyperbolic tangent, tanh), $\bv\in\R^{T-h+1}$ is a bias vector, and $\ast$ denotes the convolutional operator.
We then apply a max-over-time pooling operation~\cite{collobert2011natural} to the feature map and take its maximum value, {\em i.e.}, $\hat{c}=\max\{\cv\}$, as the feature corresponding to this particular filter.
Convolving the same filter with the $h$-gram at every position in the sentence allows features to be extracted independently of their position in the sentence.
This pooling scheme tries to capture the most salient feature, \emph{i.e.}, the one with the highest value, for each feature map, effectively filtering out less informative compositions of words.
Further, this pooling scheme also guarantees that the extracted features are independent of the length of the input sentence.

The above process describes how one feature is extracted from one filter.
In practice, the model uses multiple filters with varying window sizes.
Each filter can be considered as a \emph{linguistic feature} detector that learns to recognize a specific class of $h$-grams.
Assume we have $m$ window sizes, and for each window size, we use $p$ filters, then we obtain a $mp$-dimensional vector $\fv$ to represent a sentence.
On top of this $mp$-dimensional feature vector, we specify a softmax layer to map the input sentence to an output $D(\Xmat) \in [0,1]$, representing the probability of $\Xmat$ being from the data distribution (real), rather than from the adversarial generator (synthesized).

There are other CNN architectures in the literature~\cite{kalchbrenner2014convolutional,hu2014convolutional,johnson2014effective}.
We adopt the CNN model of \citet{kim2014convolutional,collobert2011natural} due to its simplicity and excellent performance on sentence classification tasks.

\vspace{-2.0mm}
\paragraph{LSTM generator}
We specify an LSTM generator to translate a latent code vector, $\zv$, into a synthetic sentence $\tilde{s}$.
This is illustrated in Figure~\ref{fig:framework}(bottom).
The probability of a length-$T$ sentence, $\tilde{s}$, given the encoded feature vector, $\zv$, is defined as
\begin{align}\label{eq:lik}
p(\tilde{s}|\zv)=p(\tilde{w}^1|\zv)\prod_{t=2}^{T}p(\tilde{w}^t|\tilde{w}^{<t},\zv) \,,
\end{align}
where $\tilde{w}^t$ denotes the $t$-th generated token.
Specifically, we generate the first word $\tilde{w}^1$, deterministically from $\zv$, with $ (\tilde{w}^1|\zv) = \mbox{argmax}(\Vmat\hv_1)$, where $\hv_1=\tanh(\Cmat \zv)$.
Bias terms are omitted for simplicity.
All other words in the sentence are sequentially generated using the RNN, based on previously generated words, until the end-sentence symbol is generated.
The $t$-th word $ \tilde{w}^t$ is generated as $(\tilde{w}^t|\tilde{w}^{<t},\zv) = \mbox{argmax} (\Vmat \hv_t)$, where $<t \triangleq \{1,\ldots,t-1\}$, and the hidden units $\hv_t$ are recursively updated through $\hv_t = \Ucal(\yv_{t-1},\hv_{t-1},\zv)$. $\Vmat$ is a weight matrix used for computing a distribution over words.
The input $\yv_{t-1}$ for the $t$-th step is the embedding vector of the previous generated word $\tilde{w}^{t-1}$, \ie,
\begin{align}\label{eq:argmax}
\yv_{t-1} =\Wmat_e[\tilde{w}^{t-1}] \,.
\end{align}
The synthetic sentence $\tilde{s} = [\tilde{w}^{1},\cdots,\tilde{w}^{L}]$ is deterministically obtained given $\zv$ by concatenating the generated words.
In experiments, the transition function, $\Ucal(\cdot)$, is implemented with an LSTM~\cite{hochreiter1997long}.
Details are provided in the Supplementary Material.

\subsection{Training Techniques}
\paragraph{Soft-argmax approximation}
To train the generator $G(\cdot)$, which contains discrete variables, direct application of the gradient estimation may be difficult \cite{yu2016sequence}.
Score-function-based approaches, such as the REINFORCE algorithm \cite{williams1992simple}, achieve unbiased gradient estimation for discrete variables using Monte Carlo estimation.
However, in our experiments, we found that the variance of the gradient estimation is very large, which is consistent with \citet{maddison2016concrete}.
Here we consider a \emph{soft-argmax} operator~\cite{zhanggenerating}, i.e., when performing learning, as an approximation to \eqref{eq:argmax}:
\begin{align}\label{eq:softargmax}
\yv_{t-1} =\Wmat_e \text{softmax}(\Vmat \hv_{t-1} \odot L ) \,.
\end{align}
where $\odot$ represents the element-wise product.
Note that when $L\rightarrow \infty$, this approximation approaches \eqref{eq:argmax}.

\paragraph{Pre-training}
Previous literature \cite{goodfellow2014generative,salimans2016improved} has discussed the fundamental difficulty of training GANs using gradient-based methods.
In general, gradient descent optimization schemes may fail to converge to the equilibrium by moving along the orbit trajectory among saddle points \cite{salimans2016improved}.
Intuitively, good initialization can facilitate convergence.
Toward this end, we initialize the LSTM parameters of the generator by pre-training a standard CNN-LSTM autoencoder~\cite{gan2016unsupervised}.
For the discriminator/encoder initialization, we use a {\it permutation training} strategy.
For each sentence in the corpus, we randomly swap two words to construct a slightly \emph{tweaked} sentence counterpart.
The discriminator is pre-trained to distinguish the tweaked sentences from the true sentences.
The swapping operation is preferred here because it constitutes a much more challenging task for the discriminator to learn, compared to adding or deleting words, where the structure of real sentences is more strongly disrupted, thus making it easier for the discriminator. The permutation pre-training is important because it requires the discriminator to learn features characteristic of sentences' long dependencies. We empirically found this provides a better initialization (compared to no pre-training) for the discriminator to learn good features. 

We also utilized other training techniques to stabilize training, such as soft-labeling \cite{salimans2016improved}.
Details of these are provided in the Supplementary Material.

\section{Related Work}\label{sec:related}
Generative Moment Matching Networks (GMMNs) \cite{dziugaite2015training,li2015generative} are closely related to our approach.
However, these methods either directly match the empirical distribution in the data domain, or extract features using a pre-trained autoencoder \cite{li2015generative}.
If the goal is to perform matching in the data domain when generating sentences, the dimensionality of input data would be $T\times k$ (higher than 10,000 in our case).
Note that the minibatch size required to obtain reasonable statistical power grows linearly with the number of dimension \cite{ramdas2014high}, and the computational cost of MMD grows quadratically with the size of data points.
Therefore, directly applying GMMNs is often computationally prohibitive.
Furthermore, directly matching in the data domain via GMMNs implies word-by-word discrepancy, which yields less smooth gradients.
This happens because a word-by-word discrepancy ignores sentence structure.
For example, two sentences ``a boy is swimming'' and ``boy is swimming'' will be far apart in a word-by-word metric, when they are indeed close in a sentence-by-sentence feature space.

A two-step method, where a feature encoder is generated first as in \citet{li2015generative} helps alleviate the problems above.
However, in \citet{li2015generative} the feature encoder is fixed once pre-trained, limiting the potential to adjust features during the training phase.
Alternatively, our approach matches the real and synthetic data on a sentence feature space, where features are \emph{dynamically} and \emph{adversarially} adapted to focus on the most challenging features for the generator to mimic.
In addition, features are designed to maintain both discrimination and reconstruction ability, instead of merely focusing on reconstruction as in \citet{li2015generative}.

Recent work considered combining autoencoders or variational autoencoders \cite{kingma2013auto} with GAN \cite{zhao:2016wu,larsen:2015vi,makhzani:2015tm,mescheder:2017vp,wang2016learning}.
They demonstrated superior performance on image generation.
Our approach is similar to these approaches; however, we attempt to learn the reconstruction of the latent code, instead of the input data (sentences).
\citet{Donahue:2016wo,dumoulin2016adversarially} learned a reverse mapping from data space to latent space.
In our approach we enforce the discriminator and encoder to share a latent structure, with the aim of learning a representation for both discrimination and latent code reconstruction.
\citet{chen2016infogan} maximized the mutual information between the generated data and the latent codes by leveraging a network-adapted variational proposal distribution.
In our case, we minimize the distance between the original and reconstructed latent codes.

Our approach attempts to minimize a NN-based embedded MMD distance of two empirical distributions.
Aside from MMD, kernel-based discrepancy metrics such as kernelized Stein discrepancy \cite{liu2016kernelized,wang2016learning} have been shown to be computationally tractable, while maintaining statistical power.
We leave the investigation of using Stein for moment matching as a promising future direction.
Wasserstein GAN \cite{arjovsky2017wasserstein} considers an Earth-Mover (EM) distance of the real data and synthetic data distribution, instead of the JSD as in standard GAN \cite{goodfellow2014generative} or TVD as in \citet{zhao:2016wu}.
The EM metric yields stable gradients, thus avoiding the collapsing mode and vanishing gradient problem of the latter two.
We note that our approach is equivalent to minimizing a MMD loss over the data domain, however, with a NN-based embedded Gaussian kernel.
As shown in \citet{arjovsky2017wasserstein}, MMD is a proper metric when the kernel is universal.
Because of the similarity of the conditions, 
our approach enjoys the advantages of Wasserstein GAN, namely, ameliorating the gradient vanishing problems.

\section{Experiments}
\paragraph{Data and Experimental Setup}
Our model is trained using a combination of two datasets: (\emph{i}) the BookCorpus dataset~\cite{zhu2015aligning}, which consists of 70 million sentences from over 7000 books; and (\emph{ii}) the ArXiv dataset, which consists of 5 million sentences from abstracts of papers from various subjects, obtained from the arXiv website.
The motivation for merging two different corpora is to investigate whether the model can generate sentences that integrate both scientific and informal writing styles.
We randomly choose 0.5 million sentences from BookCorpus and 0.5 million sentences from arXiv to construct training and validation sets, \ie, 1 million sentences for each.
For testing, we randomly select 25,000 sentences from both corpus, for a total of 50,000 sentences.

We train the generator and discriminator/encoder iteratively.
Provided that the LSTM generator typically involves more parameters and is more difficult to train than the CNN discriminator, we perform one optimization step for the discriminator for every $K=5$ steps of the generator.
We use a mixture of 5 isotropic Gaussian (RBF) kernels with different bandwidths $\sigma$ as in \citet{li2015generative}.
Bandwidth parameters are selected to be close to the median distance (in our case around 20) of feature vectors encoded from real sentences.
$\lambda_{r}$ and $\lambda_{m}$ are selected based on the performance on the validation set.
The validation performance is evaluated by loss of generator and corpus-level BLEU score \cite{papineni2002bleu}, described below.

For the CNN discriminator/encoder, we use filter windows ($h$) of sizes \{3,4,5\} with 300 feature maps each, hence each sentence is represented as a 900-dimensional vector. The dimensionality of $\zv$ and $\hat{\zv}$ is also 900.
The feature vector is then fed into a 900-200-2 fully connected network for the discriminator and 900-900-900 for encoder, with sigmoid activation units connecting the intermediate layers and softmax/tanh units for the top layer of discriminator/encoder.
We did not observe performance changes by adding dropout.
For the LSTM sentence generator, we use one hidden layer of 500 units.

Gradients are clipped if the norm of the parameter vector exceeds 5~\cite{sutskever2014sequence}.
Adam \cite{kingma2014adam} with learning rate $5\times10^{-5}$ for both discriminator and generator is utilized for optimization. The size of the minibatch is set to 256.

Both the generator and the discriminator are pre-trained using the strategies described in Section~\ref{sec:model}.
We also employed a \emph{warm-up} training during the first two epochs, as we found it improves convergence during the initial stage of learning.
Specifically, we use a mean-matching objective for the generator loss, \ie, $||\mathbb{E} \fv - \mathbb{E} \tilde{\fv}||^2$, as in \citet{salimans2016improved}.
Further details of the experimental design are provided in the the Supplementary Material.
All experiments are implemented in Theano~\cite{bastien2012theano}, using one NVIDIA GeForce GTX TITAN X GPU with 12GB memory.
The model was trained for 50 epochs in roughly 3 days.
Learning curves are shown in the Supplementary Material.

\begin{figure}[t!]
	\centering
	\includegraphics[width=.23\textwidth]{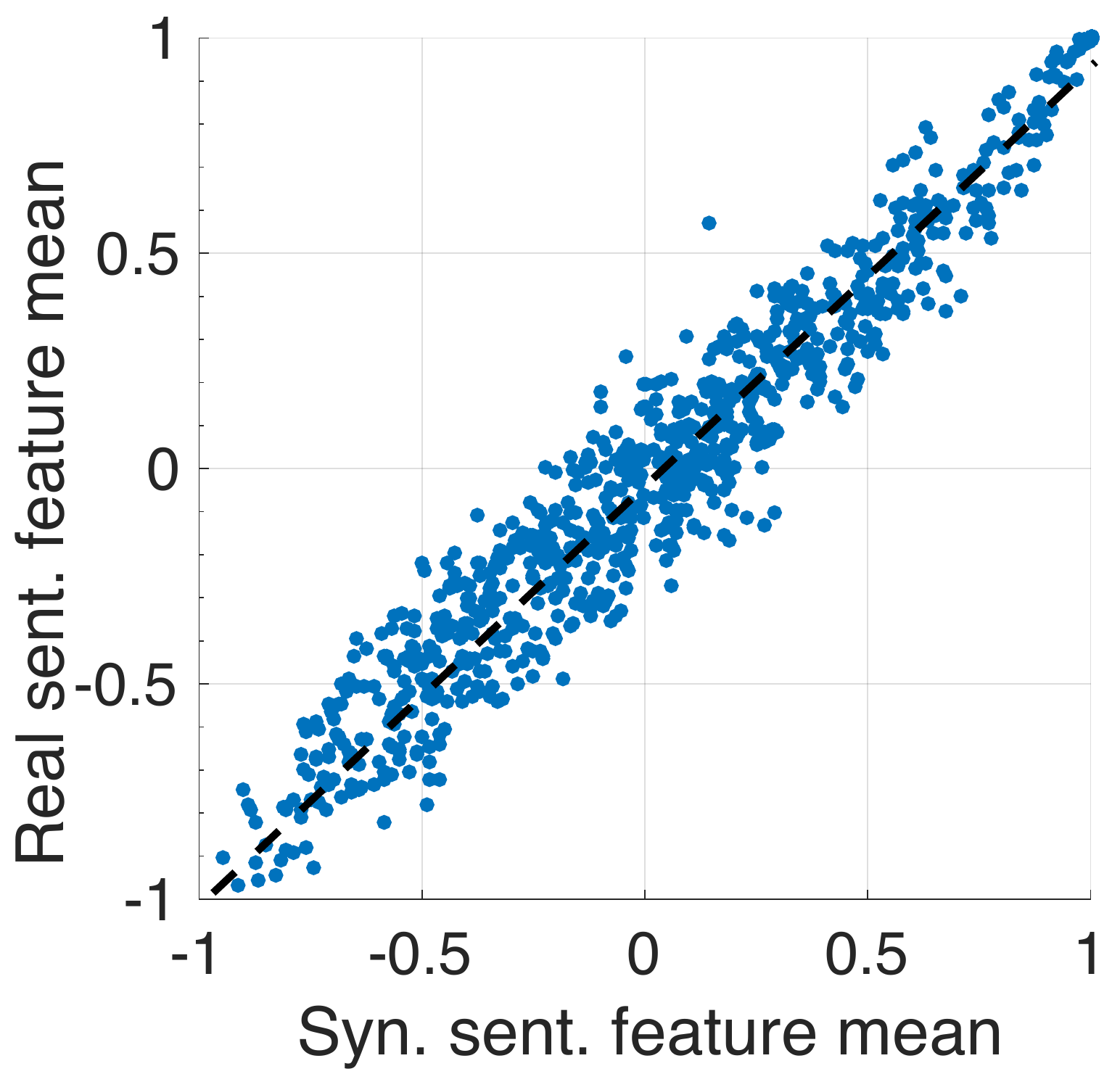}
	\includegraphics[width=.23\textwidth]{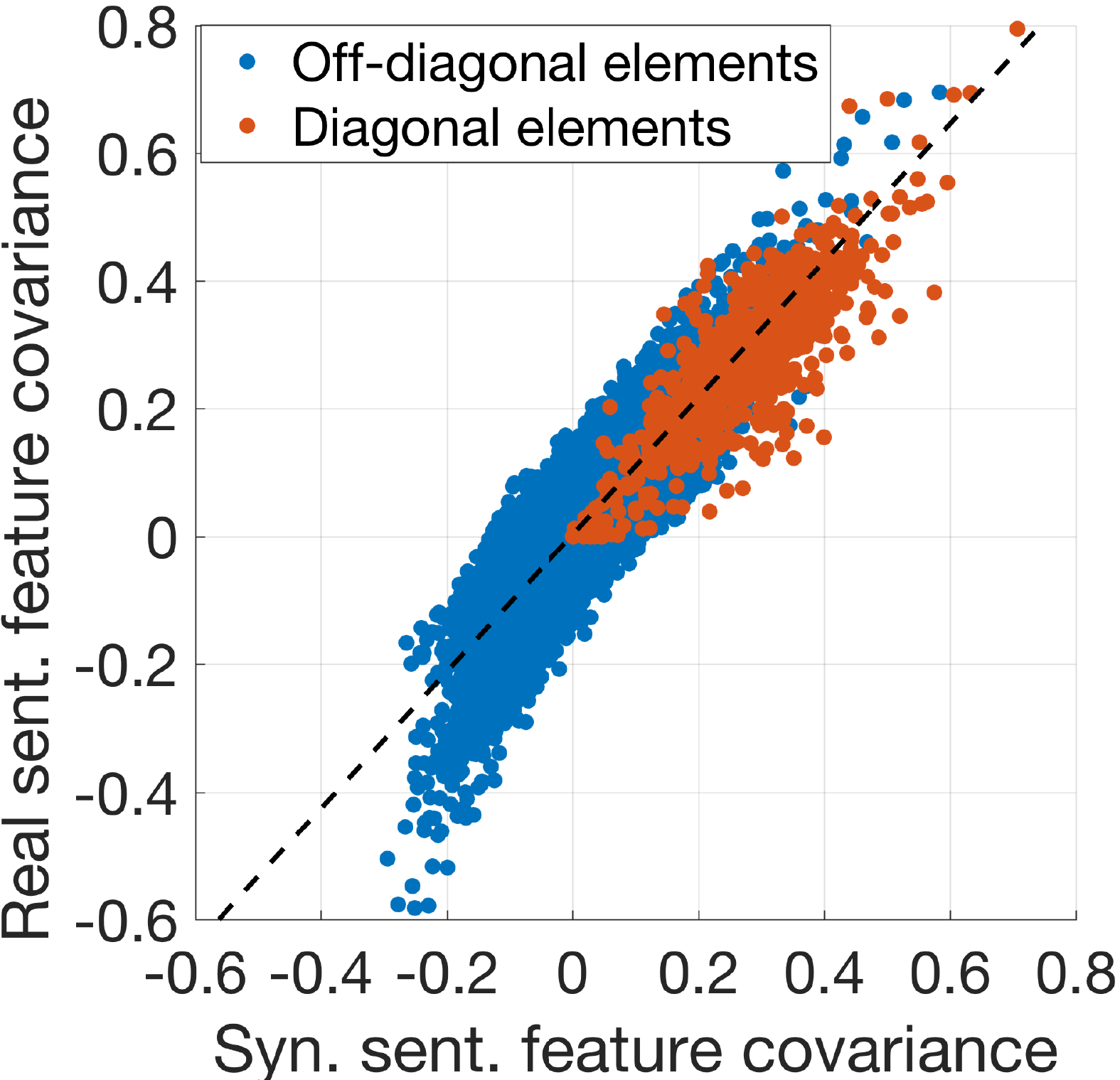}
	\caption{Moment matching comparison. Left: expectations of latent features from real \emph{vs.} synthetic data. Right: elements of $\tilde{\Sigmamat}_{i,j,\fv}$ \emph{vs.} $\tilde{\Sigmamat}_{i,j,\tilde{\fv}}$, for real and synthetic data, respectively. 
	}
	\label{fig:figs_feature_matching}
	\vspace{-3.5mm}
\end{figure}

\paragraph{Matching feature distributions}
We first examine the generator's ability to produce synthetic features similar to those obtained from real data.
For this purpose, we calculate the empirical expectation of the 900-dimensional sentence feature vector over 2,000 real sentences and 2,000 synthetic sentences.
As shown in Figure~\ref{fig:figs_feature_matching}(left), the expectation of these 900 feature dimensions from synthetic sentences matches well with the feature expectation from the real sentences.
We also compared the estimated covariance matrix elements $\tilde{\Sigmamat}_{i,j,\fv}$ (including $900*899/2$ off-diagonal elements and $900$ diagonal elements) from real data against the covariance matrix elements $\tilde{\Sigmamat}_{i,j,\tilde{\fv}}$ estimated from synthetic data, in Figure~\ref{fig:figs_feature_matching}(right).
We observe that the covariance structure of the 900-dimensional features from real and synthetic sentences in general match well.
The full covariance matrices for real and synthetic sentences are provided in the Supplementary Material.
We observe that the (mapped) synthetic features nicely cover the real sentence features density, while ``completing'' other areas of low density. 

\vspace{-2.5mm}
\paragraph{Quantitative comparison}
We evaluate the generated-sentence quality using the BLEU score \cite{papineni2002bleu} and Kernel Density Estimation (KDE), as in \citet{goodfellow2014generative,nowozin2016f}.
For comparison, we consider textGAN with 4 different loss objectives: Mean Matching (MM) as in \citet{salimans2016improved}, Covariance Matching (CM) as in \eqref{eq:cov_match}, MMD and MMD with compressed network (MMD-L), by mapping the original 900-dimensional features to 200-dimensional, as described in Section~\ref{sec:alter}.
We also compare to a baseline autoencoder (AE) model.
The AE uses a CNN as encoder and an LSTM as decoder, where the CNN and LSTM network structures are set to be identical as the CNN and LSTM used in textGAN.
We finally consider a Variational Autoencoder (VAE) implemented as in \citet{bowman2015generating}.
To train the VAE model, we use annealing to gradually increase the KL divergence between the prior and approximated posterior.
The details are provided in the the Supplementary Material.
We also compare with seqGAN \cite{yu2016sequence}.
For seqGAN we follow the authors' guidelines of running 350 pre-training epochs followed by 50 discriminator training epochs, to generate 320 sentences.
For AE, VAE and textGAN, we first uniformly sample 320 latent codes from the latent code space, and use the corresponding generator (or decoder, in the AE/VAE case) to generate sentences.

\begin{table}[t!]
	\caption{Quantitative results using BLEU-2,3,4 and KDE.}\label{tab:evaluation}
	\vskip 0.1in
	\begin{scriptsize}
		\begin{tabular}{ccccc}
			\hline
			&BLEU-4 &BLEU-3 &BLEU-2 &KDE(nats)\\
			\hline
			AE 		& 0.01$\pm$0.01 & 0.11$\pm$0.02 & 0.39$\pm$0.02 & 2727$\pm$42\\
			VAE 			& 0.02$\pm$0.02 & 0.16$\pm$0.03 & 0.54$\pm$0.03 & 1892$\pm$25\\
			seqGAN 			& 0.04$\pm$0.04 & 0.30$\pm$0.08 & 0.67$\pm$0.04 & 2019$\pm$53\\
			textGAN(MM) 	& 0.09$\pm$0.04 & 0.42$\pm$0.04 & 0.77$\pm$0.03 & 1823$\pm$50 \\
			textGAN(CM) 	& 0.12$\pm$0.03 & 0.49$\pm$0.06 & 0.84$\pm$0.02 & 1686$\pm$41\\
			textGAN(MMD) 	& {\bf 0.13$\pm$0.05} & 0.49$\pm$0.06 & 0.83$\pm$0.04 & 1688$\pm$38\\
			textGAN(MMD-L) 	& 0.11$\pm$0.05 & {\bf0.52$\pm$0.07} & {\bf 0.85$\pm$0.04} & {\bf 1684$\pm$44}\\
			\hline
		\end{tabular}
	\end{scriptsize}
	\vspace{-2.5mm}
\end{table}

For BLEU score evaluation, we follow the strategy in \citet{yu2016sequence} of using the entire test set as the reference.
For KDE evaluation, the lengths of the generated sentences are different, thus we first embed all the sentences to a 900-dimensional vector.
Since no standard sentence encoder is available, we use the encoder learned from AE.
The covariance matrix for the Parzen kernel in KDE is set to be the covariance of feature vectors for real tested sentences.
Despite the fact that the KDE approach, as a log-likelihood estimator tends to have high variance \cite{theis2015note}, the KDE score tracks well with our BLEU score evaluation.

The results are shown in Table~\ref{tab:evaluation}.
MMD and MMD-L generally score higher in sentences quality.
MMD-L seems better at capturing 2-grams (BLEU-2), while MMD outperforms MMD-L in 4-grams (BLEU-4).
We also observed that when using CM, the generated sentences tend to be shorter than MMD (not shown).
%

\paragraph{Generated sentences}
Table~\ref{tab:sentence_generation} shows six sentences generated by textGAN.
Note that the generated sentences seem to be able to produce novel phrases by imagining concept combinations, \eg, in Table~\ref{tab:sentence_generation}(b,c,f), or to borrow words from a different corpus to compose novel sentences, \eg, in Table~\ref{tab:sentence_generation}(d,e).
In many cases, it learns to automatically match the parentheses and quotation marks, \eg, in Table~\ref{tab:sentence_generation}(a), and can synthesize relatively long sentences, \eg, in \ref{tab:sentence_generation}(a,f).
In general, the synthetic sentences seem syntactically reasonable.
However, the semantic meaning is less well preserved especially in sentence of more than 20 words, \eg, in Table~\ref{tab:sentence_generation}(e,f).
\begin{table}[t!]
	\caption{Sentences generated by textGAN.}\label{tab:sentence_generation}
	\vskip 0.1in
	\begin{scriptsize}
		\begin{tabular}{l|l}
			\toprule
			\textbf{a} & we show the joint likelihood estimator ( in a large number of estimating \\ 
			&  variables embedded on the subspace learning ) .\\
			\textbf{b} & this problem achieves less interesting choices of convergence  guarantees\\
			&  on turing machine learning .\\
			\textbf{c} & in hidden markov relational spaces , the random walk feature \\
			& decomposition is unique generalized parametric mappings. \\
			\textbf{d} & i see those primitives specifying a deterministic probabilistic machine  \\ 
			&  learning algorithm . \\
			\textbf{e} & i wanted in alone in a gene expression dataset which do n't form phantom \\
			&  action values . \\			
			\textbf{f} & as opposite to a set of fuzzy modelling algorithm , pruning is performed  \\
			&  using a template representing network structures .	\\
			\bottomrule
		\end{tabular}
	\end{scriptsize}
	\vspace{-2.5mm}
\end{table}

We observe that the discriminator can still sufficiently distinguish the synthetic sentences from the real ones (the probability to predict synthetic data as real is around 0.05), even when the synthetic sentences seems to perserve reasonable grammatical structure and use proper wording.
It is likely that the CNN is able to accurately characterize the semantic meaning and differentiate sentences, while the generator may get trapped into a local optimum, where any slight modification would result in a higher loss \eqref{eq:g_step_mmd} for the generator.
Presumably, long-range distance features are not difficult to abstract by the discriminator/encoder, however, is less likely to be imitated by the generator.
One promising direction is to leverage reinforcement learning strategies as in \citet{yu2016sequence}, where the updating for LSTM can be more effectively steered.
Nevertheless, investigation on how to improve the the long-range behavior is left as interesting future work.

\paragraph{Latent feature space trajectories}
Following \citet{bowman2015generating}, we further empirically evaluate whether the latent variable space can ``densely'' encode sentences.
We visualize the transition from one sentence to another by constructing a linear path between two randomly selected points in latent feature space, to then generate the intermediate sentences along the linear trajectory.
For comparison, a baseline autoencoder (AE) is trained for 20 epochs.
The results for textGAN and AE are presented in Table~\ref{tab:sentence_transition}.
Compared to AE, the sentences produced by textGAN are generally more syntactically and semantically reasonable.
The transition suggest ``smoothness'' and interpretability, however, the wording choices and sentence structure showed dramatic changes in some regions in the latent feature space.
This seems to indicate that local ``transition smoothness'' varies from region to region.

\begin{table}[t!]
	\caption{Intermediate sentences produced from linear transition between two points (A and B) in the latent feature space. Each sentence is generated from a latent point on a linear path.}\label{tab:sentence_transition}
	\centering
	\vskip 0.1in
	\begin{scriptsize}
		\begin{tabular}{lp{1.3in}|p{1.35in}}
			& \multicolumn{1}{c}{\bf textGAN} & \multicolumn{1}{c}{\bf AE} \\
			\hline
			\textbf{A} & \multicolumn{2}{c}{ our methods apply novel approaches to solve modeling tasks .} \\
			\hline
			\textbf{-} & our methods apply novel approaches to solve modeling . & our methods apply to train UNK models involving complex .\\
			\hline
			\textbf{-} & our methods apply two different approaches to solve computing . & our methods solve use to train ) . \\
			\hline
			\textbf{-} & our methods achieves some different approaches to solve computing . & our approach show UNK to models exist .\\
			\hline
			\textbf{-} & our methods achieves the best expert structure detection . & that supervised algorithms show to UNK speed . \\
			\hline
			\textbf{-} & the methods have been different related tasks . & that address algorithms to handle ) .\\
			\hline
			\textbf{-} & the guy is the minimum of UNK . & that address versions to be used in .\\
			\hline
			\textbf{-} & the guy is n't easy tonight . & i believe the means of this attempt to cope .\\
			\hline
			\textbf{-} & i believe the guy is n't smart okay? & i believe it 's we be used to get .\\
			\hline
			\textbf{-} & i believe the guy is n't smart . & i believe it i 'm a way to belong .\\
			\hline
			\textbf{B} & \multicolumn{2}{c}{i believe i 'm going to get out .} \\
			\hline
		\end{tabular}
	\end{scriptsize}
	\vspace{-1.5mm}
\end{table}

\section{Conclusion}
We have introduced a novel approach for text generation using adversarial training, termed TextGAN, and have discussed several techniques to specify and train such a model.
We demonstrated that the proposed model delivers superior performance compared to related approaches, can produce realistic sentences, and that the learned latent representation space can ``smoothly'' encode plausible sentences. We quantitatively evaluate the proposed methods with baseline models and existing methods. The results indicate superior performance of TextGAN.

In future work, we will attempt to apply conditional GAN models \cite{mirza2014conditional} to disentangle the latent representations for different writing styles.
This would enable a smooth lexical and grammatical transition between different writing styles.
It would be also interesting to generate text by conditioning on observed images \cite{pu2016variational}.
In addition, we plan to leverage an additional refining stage where a reverse-order LSTM \cite{graves2005framewise} is applied after the sentence is first generated, to produce sentences with better long-term semantical interpretation.

\section*{Acknowledgments} 
This research was supported by ARO, DARPA, DOE, NGA, ONR and NSF.
\clearpage
{\small
	\bibliographystyle{icml2017}
	\bibliography{references}

\begin{thebibliography}{47}
\providecommand{\natexlab}[1]{#1}
\providecommand{\url}[1]{\texttt{#1}}
\expandafter\ifx\csname urlstyle\endcsname\relax
  \providecommand{\doi}[1]{doi: #1}\else
  \providecommand{\doi}{doi: \begingroup \urlstyle{rm}\Url}\fi

\bibitem[Arjovsky \& Bottou(2017)Arjovsky and Bottou]{arjovsky2017towards}
Arjovsky, Martin and Bottou, L{\'e}on.
\newblock Towards principled methods for training generative adversarial
  networks.
\newblock In \emph{ICLR}, 2017.

\bibitem[Arjovsky et~al.(2017)Arjovsky, Chintala, and
  Bottou]{arjovsky2017wasserstein}
Arjovsky, Martin, Chintala, Soumith, and Bottou, L{\'e}on.
\newblock Wasserstein gan.
\newblock In \emph{ICML}, 2017.

\bibitem[Bastien et~al.(2012)Bastien, Lamblin, Pascanu, Bergstra, Goodfellow,
  Bergeron, Bouchard, Warde-Farley, and Bengio]{bastien2012theano}
Bastien, F., Lamblin, P., Pascanu, R., Bergstra, J., Goodfellow, I., Bergeron,
  A., Bouchard, N., Warde-Farley, D., and Bengio, Y.
\newblock Theano: new features and speed improvements.
\newblock \emph{arXiv:1211.5590}, 2012.

\bibitem[Bengio et~al.(2015)Bengio, Vinyals, Jaitly, and
  Shazeer]{bengio2015scheduled}
Bengio, Samy, Vinyals, Oriol, Jaitly, Navdeep, and Shazeer, Noam.
\newblock Scheduled sampling for sequence prediction with recurrent neural
  networks.
\newblock In \emph{NIPS}, 2015.

\bibitem[Bowman et~al.(2016)Bowman, Vilnis, Vinyals, Dai, Jozefowicz, and
  Bengio]{bowman2015generating}
Bowman, Samuel~R, Vilnis, Luke, Vinyals, Oriol, Dai, Andrew~M, Jozefowicz,
  Rafal, and Bengio, Samy.
\newblock Generating sentences from a continuous space.
\newblock In \emph{CoNLL}, 2016.

\bibitem[Chen et~al.(2016)Chen, Duan, Houthooft, Schulman, Sutskever, and
  Abbeel]{chen2016infogan}
Chen, Xi, Duan, Yan, Houthooft, Rein, Schulman, John, Sutskever, Ilya, and
  Abbeel, Pieter.
\newblock Infogan: Interpretable representation learning by information
  maximizing generative adversarial nets.
\newblock In \emph{NIPS}, 2016.

\bibitem[Cho et~al.(2014)Cho, Van~Merri{\"e}nboer, Gulcehre, Bahdanau,
  Bougares, Schwenk, and Bengio]{cho2014learning}
Cho, K., Van~Merri{\"e}nboer, B., Gulcehre, C., Bahdanau, D., Bougares, F.,
  Schwenk, H., and Bengio, Y.
\newblock Learning phrase representations using rnn encoder-decoder for
  statistical machine translation.
\newblock In \emph{EMNLP}, 2014.

\bibitem[Collobert et~al.(2011)Collobert, Weston, Bottou, Karlen, Kavukcuoglu,
  and Kuksa]{collobert2011natural}
Collobert, R., Weston, J., Bottou, L., Karlen, M., Kavukcuoglu, K., and Kuksa,
  P.
\newblock Natural language processing (almost) from scratch.
\newblock In \emph{JMLR}, 2011.

\bibitem[Donahue et~al.(2017)Donahue, Kr{\"a}henb{\"u}hl, and
  Darrell]{Donahue:2016wo}
Donahue, Jeff, Kr{\"a}henb{\"u}hl, Philipp, and Darrell, Trevor.
\newblock Adversarial feature learning.
\newblock In \emph{ICLR}, 2017.

\bibitem[Dumoulin et~al.(2016)Dumoulin, Belghazi, Poole, Lamb, Arjovsky,
  Mastropietro, and Courville]{dumoulin2016adversarially}
Dumoulin, Vincent, Belghazi, Ishmael, Poole, Ben, Lamb, Alex, Arjovsky, Martin,
  Mastropietro, Olivier, and Courville, Aaron.
\newblock Adversarially learned inference.
\newblock \emph{arXiv preprint arXiv:1606.00704}, 2016.

\bibitem[Dziugaite et~al.(2015)Dziugaite, Roy, and
  Ghahramani]{dziugaite2015training}
Dziugaite, Gintare~Karolina, Roy, Daniel~M, and Ghahramani, Zoubin.
\newblock Training generative neural networks via maximum mean discrepancy
  optimization.
\newblock \emph{arXiv:1505.03906}, 2015.

\bibitem[Gan et~al.(2016)Gan, Pu, Henao, Li, He, and
  Carin]{gan2016unsupervised}
Gan, Zhe, Pu, Yunchen, Henao, Ricardo, Li, Chunyuan, He, Xiaodong, and Carin,
  Lawrence.
\newblock Unsupervised learning of sentence representations using convolutional
  neural networks.
\newblock \emph{arXiv preprint arXiv:1611.07897}, 2016.

\bibitem[Goodfellow et~al.(2014)Goodfellow, Pouget-Abadie, Mirza, Xu,
  Warde-Farley, Ozair, Courville, and Bengio]{goodfellow2014generative}
Goodfellow, Ian, Pouget-Abadie, Jean, Mirza, Mehdi, Xu, Bing, Warde-Farley,
  David, Ozair, Sherjil, Courville, Aaron, and Bengio, Yoshua.
\newblock Generative adversarial nets.
\newblock In \emph{NIPS}, 2014.

\bibitem[Graves \& Schmidhuber(2005)Graves and
  Schmidhuber]{graves2005framewise}
Graves, Alex and Schmidhuber, J{\"u}rgen.
\newblock Framewise phoneme classification with bidirectional lstm and other
  neural network architectures.
\newblock \emph{Neural Networks}, 2005.

\bibitem[Gretton et~al.(2012)Gretton, Borgwardt, Rasch, Sch{\"o}lkopf, and
  Smola]{gretton2012kernel}
Gretton, Arthur, Borgwardt, Karsten~M, Rasch, Malte~J, Sch{\"o}lkopf, Bernhard,
  and Smola, Alexander.
\newblock A kernel two-sample test.
\newblock \emph{JMLR}, 2012.

\bibitem[Hochreiter \& Schmidhuber(1997)Hochreiter and
  Schmidhuber]{hochreiter1997long}
Hochreiter, S. and Schmidhuber, J.
\newblock Long short-term memory.
\newblock In \emph{Neural computation}, 1997.

\bibitem[Hu et~al.(2014)Hu, Lu, Li, and Chen]{hu2014convolutional}
Hu, B., Lu, Z., Li, H., and Chen, Q.
\newblock Convolutional neural network architectures for matching natural
  language sentences.
\newblock In \emph{NIPS}, 2014.

\bibitem[Husz{\'a}r(2015)]{huszar2015not}
Husz{\'a}r, Ferenc.
\newblock How (not) to train your generative model: Scheduled sampling,
  likelihood, adversary?
\newblock \emph{arXiv:1511.05101}, 2015.

\bibitem[Johnson \& Zhang(2015)Johnson and Zhang]{johnson2014effective}
Johnson, R. and Zhang, T.
\newblock Effective use of word order for text categorization with
  convolutional neural networks.
\newblock In \emph{NAACL HLT}, 2015.

\bibitem[Kalchbrenner et~al.(2014)Kalchbrenner, Grefenstette, and
  Blunsom]{kalchbrenner2014convolutional}
Kalchbrenner, N., Grefenstette, E., and Blunsom, P.
\newblock A convolutional neural network for modelling sentences.
\newblock In \emph{ACL}, 2014.

\bibitem[Kim(2014)]{kim2014convolutional}
Kim, Y.
\newblock Convolutional neural networks for sentence classification.
\newblock In \emph{EMNLP}, 2014.

\bibitem[Kingma \& Ba(2015)Kingma and Ba]{kingma2014adam}
Kingma, D. and Ba, J.
\newblock Adam: A method for stochastic optimization.
\newblock In \emph{ICLR}, 2015.

\bibitem[Kingma \& Welling(2014)Kingma and Welling]{kingma2013auto}
Kingma, Diederik~P and Welling, Max.
\newblock Auto-encoding variational bayes.
\newblock In \emph{ICLR}, 2014.

\bibitem[Lamb et~al.(2016)Lamb, GOYAL, Zhang, Zhang, Courville, and
  Bengio]{lamb2016professor}
Lamb, Alex~M, GOYAL, Anirudh Goyal ALIAS~PARTH, Zhang, Ying, Zhang, Saizheng,
  Courville, Aaron~C, and Bengio, Yoshua.
\newblock Professor forcing: A new algorithm for training recurrent networks.
\newblock In \emph{NIPS}, 2016.

\bibitem[Larsen et~al.(2016)Larsen, S{\o}nderby, Larochelle, and
  Winther]{larsen:2015vi}
Larsen, Anders Boesen~Lindbo, S{\o}nderby, S{\o}ren~Kaae, Larochelle, Hugo, and
  Winther, Ole.
\newblock Autoencoding beyond pixels using a learned similarity metric.
\newblock In \emph{ICML}, 2016.

\bibitem[Li et~al.(2016)Li, Monroe, Ritter, Galley, Gao, and
  Jurafsky]{li2016deep}
Li, Jiwei, Monroe, Will, Ritter, Alan, Galley, Michel, Gao, Jianfeng, and
  Jurafsky, Dan.
\newblock Deep reinforcement learning for dialogue generation.
\newblock In \emph{EMNLP}, 2016.

\bibitem[Li et~al.(2017)Li, Monroe, Shi, Ritter, and
  Jurafsky]{li2017adversarial}
Li, Jiwei, Monroe, Will, Shi, Tianlin, Ritter, Alan, and Jurafsky, Dan.
\newblock Adversarial learning for neural dialogue generation.
\newblock \emph{arXiv:1701.06547}, 2017.

\bibitem[Li et~al.(2015)Li, Swersky, and Zemel]{li2015generative}
Li, Yujia, Swersky, Kevin, and Zemel, Richard~S.
\newblock Generative moment matching networks.
\newblock In \emph{ICML}, 2015.

\bibitem[Liu et~al.(2016)Liu, Lee, and Jordan]{liu2016kernelized}
Liu, Qiang, Lee, Jason~D, and Jordan, Michael~I.
\newblock A kernelized stein discrepancy for goodness-of-fit tests.
\newblock In \emph{ICML}, 2016.

\bibitem[Maddison et~al.(2017)Maddison, Mnih, and Teh]{maddison2016concrete}
Maddison, Chris~J, Mnih, Andriy, and Teh, Yee~Whye.
\newblock The concrete distribution: A continuous relaxation of discrete random
  variables.
\newblock In \emph{ICLR}, 2017.

\bibitem[Makhzani et~al.(2015)Makhzani, Shlens, Jaitly, Goodfellow, and
  Frey]{makhzani:2015tm}
Makhzani, Alireza, Shlens, Jonathon, Jaitly, Navdeep, Goodfellow, Ian, and
  Frey, Brendan.
\newblock Adversarial autoencoders.
\newblock \emph{arXiv:1511.05644}, 2015.

\bibitem[Mescheder et~al.(2017)Mescheder, Nowozin, and
  Geiger]{mescheder:2017vp}
Mescheder, Lars, Nowozin, Sebastian, and Geiger, Andreas.
\newblock Adversarial variational bayes: Unifying variational autoencoders and
  generative adversarial networks.
\newblock In \emph{ICML}, 2017.

\bibitem[Metz et~al.(2017)Metz, Poole, Pfau, and
  Sohl-Dickstein]{metz2016unrolled}
Metz, Luke, Poole, Ben, Pfau, David, and Sohl-Dickstein, Jascha.
\newblock Unrolled generative adversarial networks.
\newblock In \emph{ICLR}, 2017.

\bibitem[Mirza \& Osindero(2014)Mirza and Osindero]{mirza2014conditional}
Mirza, Mehdi and Osindero, Simon.
\newblock Conditional generative adversarial nets.
\newblock \emph{arXiv:1411.1784}, 2014.

\bibitem[Nowozin et~al.(2016)Nowozin, Cseke, and Tomioka]{nowozin2016f}
Nowozin, Sebastian, Cseke, Botond, and Tomioka, Ryota.
\newblock f-gan: Training generative neural samplers using variational
  divergence minimization.
\newblock In \emph{NIPS}, 2016.

\bibitem[Papineni et~al.(2002)Papineni, Roukos, Ward, and
  Zhu]{papineni2002bleu}
Papineni, Kishore, Roukos, Salim, Ward, Todd, and Zhu, Wei-Jing.
\newblock Bleu: a method for automatic evaluation of machine translation.
\newblock In \emph{ACL}, 2002.

\bibitem[Pu et~al.(2016)Pu, Gan, Henao, Yuan, Li, Stevens, and
  Carin]{pu2016variational}
Pu, Yunchen, Gan, Zhe, Henao, Ricardo, Yuan, Xin, Li, Chunyuan, Stevens,
  Andrew, and Carin, Lawrence.
\newblock Variational autoencoder for deep learning of images, labels and
  captions.
\newblock In \emph{NIPS}, 2016.

\bibitem[Ramdas et~al.(2014)Ramdas, Reddi, Poczos, Singh, and
  Wasserman]{ramdas2014high}
Ramdas, Aaditya, Reddi, Sashank~J, Poczos, Barnabas, Singh, Aarti, and
  Wasserman, Larry.
\newblock On the high-dimensional power of linear-time kernel two-sample
  testing under mean-difference alternatives.
\newblock \emph{arXiv:1411.6314}, 2014.

\bibitem[Salimans et~al.(2016)Salimans, Goodfellow, Zaremba, Cheung, Radford,
  and Chen]{salimans2016improved}
Salimans, Tim, Goodfellow, Ian, Zaremba, Wojciech, Cheung, Vicki, Radford,
  Alec, and Chen, Xi.
\newblock Improved techniques for training gans.
\newblock In \emph{NIPS}, 2016.

\bibitem[Sutskever et~al.(2014)Sutskever, Vinyals, and
  Le]{sutskever2014sequence}
Sutskever, I., Vinyals, O., and Le, Q.
\newblock Sequence to sequence learning with neural networks.
\newblock In \emph{NIPS}, 2014.

\bibitem[Theis et~al.(2016)Theis, Oord, and Bethge]{theis2015note}
Theis, Lucas, Oord, A{\"a}ron van~den, and Bethge, Matthias.
\newblock A note on the evaluation of generative models.
\newblock In \emph{ICLR}, 2016.

\bibitem[Wang \& Liu(2016)Wang and Liu]{wang2016learning}
Wang, Dilin and Liu, Qiang.
\newblock Learning to draw samples: With application to amortized mle for
  generative adversarial learning.
\newblock \emph{arXiv:1611.01722}, 2016.

\bibitem[Williams(1992)]{williams1992simple}
Williams, Ronald~J.
\newblock Simple statistical gradient-following algorithms for connectionist
  reinforcement learning.
\newblock \emph{Machine learning}, 1992.

\bibitem[Yu et~al.(2017)Yu, Zhang, Wang, and Yu]{yu2016sequence}
Yu, Lantao, Zhang, Weinan, Wang, Jun, and Yu, Yong.
\newblock Seqgan: sequence generative adversarial nets with policy gradient.
\newblock In \emph{AAAI}, 2017.

\bibitem[Zhang et~al.(2016)Zhang, Gan, and Carin]{zhanggenerating}
Zhang, Yizhe, Gan, Zhe, and Carin, Lawrence.
\newblock Generating text via adversarial training.
\newblock In \emph{NIPS Workshop on Adversarial Training}, 2016.

\bibitem[Zhao et~al.(2017)Zhao, Mathieu, and LeCun]{zhao:2016wu}
Zhao, Junbo, Mathieu, Michael, and LeCun, Yann.
\newblock Energy-based generative adversarial network.
\newblock In \emph{ICLR}, 2017.

\bibitem[Zhu et~al.(2015)Zhu, Kiros, Zemel, Salakhutdinov, Urtasun, Torralba,
  and Fidler]{zhu2015aligning}
Zhu, Y., Kiros, R., Zemel, R., Salakhutdinov, R., Urtasun, R., Torralba, A.,
  and Fidler, S.
\newblock Aligning books and movies: Towards story-like visual explanations by
  watching movies and reading books.
\newblock In \emph{ICCV}, 2015.

\end{thebibliography}
}

\end{document}